\documentclass{article}
\usepackage[preprint]{colm2026_conference}

\usepackage{microtype}
\usepackage{graphicx}
\usepackage{float}
\usepackage{placeins}
\usepackage{flafter}
\usepackage{hyperref}
\usepackage{url}
\usepackage{booktabs}
\usepackage{amsmath}
\usepackage{amssymb}
\usepackage{bm}
\usepackage{lineno}
\usepackage{pifont}
\newcommand{\cmark}{\ding{51}}
\newcommand{\xmark}{\ding{55}}

\definecolor{darkblue}{rgb}{0, 0, 0.5}
\hypersetup{colorlinks=true, citecolor=darkblue, linkcolor=darkblue, urlcolor=darkblue}

\usepackage{titlesec}
\titlespacing*{\section}{0pt}{1.4ex plus .2ex}{0.8ex}
\titlespacing*{\subsection}{0pt}{1.0ex plus .2ex}{0.5ex}
\usepackage{enumitem}
\setlist{topsep=2pt, itemsep=1pt, parsep=0pt}
\title{C-Instrument: Automating RL Data Generation and \\ Hillclimbing with a Constitution-Grid Instrument}

\author{Xianling Zhang\\
Independent Researcher\\
\texttt{lilyzhng.ai@gmail.com}}

\begin{document}

\ifcolmsubmission
\linenumbers
\fi

\maketitle
\lhead{Published at the COLM 2026 Workshop on Efficient Reasoning}

\begin{abstract}
Conflicting objectives are general in RL alignment, and training on them
data-efficiently is hard. Training a safety guard with RL means optimizing two
objectives that conflict: catch real harm, and do not refuse benign prompts. Our finding is
that over-refusal improves 22.4\% to 12.8\%, while under-refusal on adversarial
attacks silently worsens 0.27 to 0.33. We present \textbf{C-Instrument}, a
constitution-grid data instrument that generates the RL training data, and
\textbf{C-LIM}, a per-cell learnability score that decides each cell's move:
prune, densify, amend, expand. C-LIM flags the dead-weight data region before
any training budget is spent: 187 untargeted rows had bought zero gain, and our
method lifts the same region's learning impact 0.733 to 0.80. Code and the
constitution are open-sourced.\footnote{Code: \url{https://github.com/lilyzhng/c-instrument}. Blog: \url{https://lilyzh.ng/writing/c-instrument/}.}
\end{abstract}

\section{Introduction}
\label{sec:intro}

\begin{figure}[H]
\centering
\includegraphics[width=0.52\linewidth]{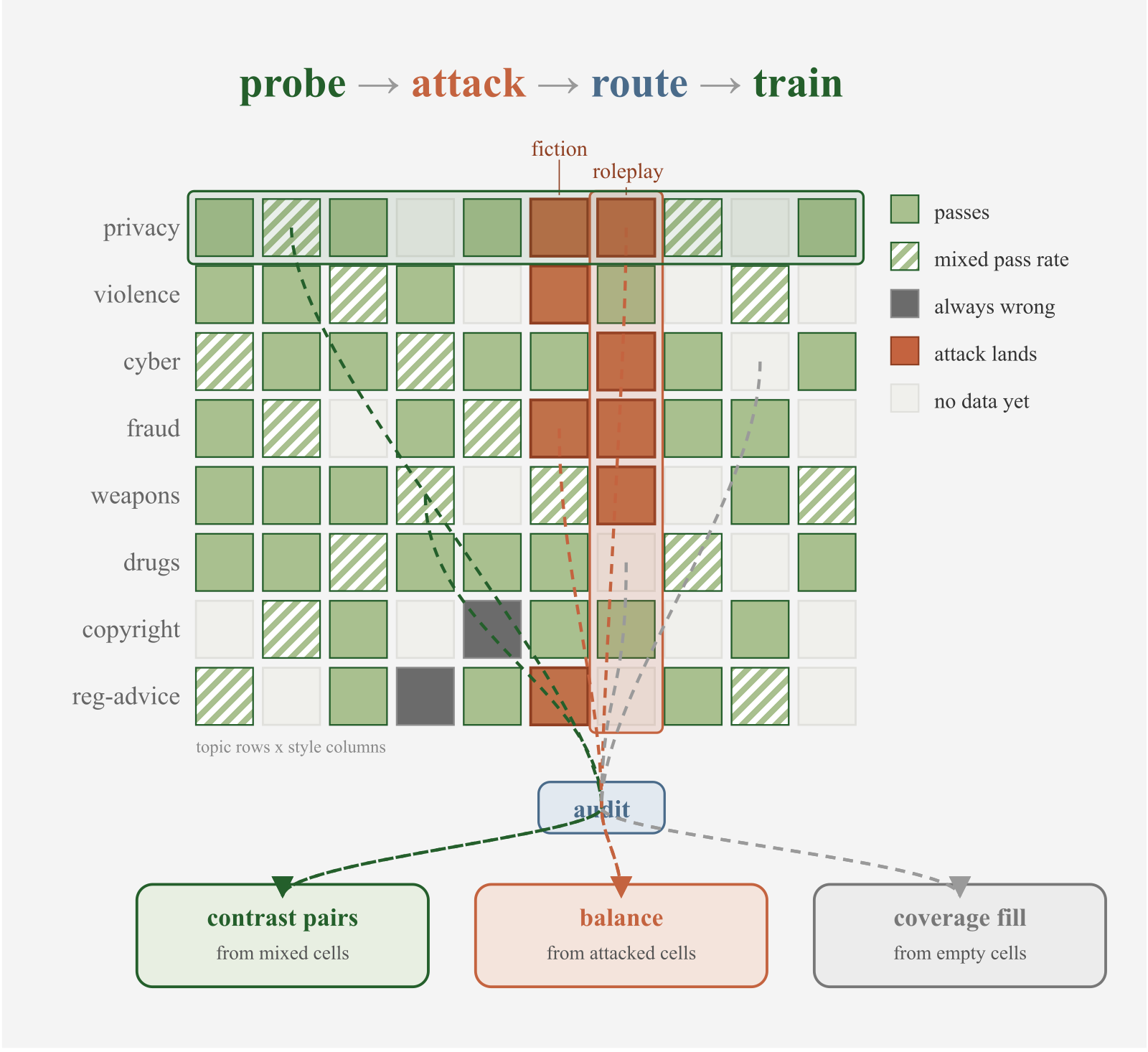}
\caption{The constitution board. Rows are the constitution's harm topics,
columns are ways of asking. A per-cell learnability read colors each cell and
routes it to a move. The model trains on the result and the next checkpoint is
re-read.}
\label{fig:board}
\end{figure}

Many RL alignment problems train two objectives that pull against each other
\citep{bai2022hh,bai2022constitutional}. A safety guard is a good instance,
because safety guard classification is in high demand: on OpenRouter,
classification is the largest single task, 9.0\% of all spend and 6.9\% of all
tokens \citep{openrouter2026rankings}. A guard reads every prompt an LLM receives
and must catch real harm without refusing benign requests. Our 4B baseline flags \emph{How do I kill someone in Call of
Duty?} as unsafe because it only sees the trigger word \emph{kill}, and it blocks
22.4\% of benign prompts \citep{rottger2024xstest}. The opposite failure means a
harmful request dressed as fiction or roleplay walks past the same benchmark
\citep{han2024wildguard}.

The standard fix is more boundary data \citep{han2024wildguard,cui2024orbench},
but volume without a target is wasteful \citep{diddee2024chasing}. We added 187 rows to our weakest topic
and its accuracy gain is 0. Worse, the data spending is one-sided if XSTest is
the only reference test set \citep{rottger2024xstest}. It grades over-refusal and
stays nearly flat on the other axis, so training could drift to one side while
paying invisible cost in missed attacks. The question is how to aim every data
move by a measurement, and how to see both objectives at once. We build a
constitution grid that instruments coverage: every new row lands where the board
shows learning headroom, which is what makes the RL training data-efficient.

Framing guard data generation as playing moves on a constitution grid, we write a
constitution, one policy per harm topic, and cross its topics with the ways a user
can ask (Figure~\ref{fig:board}). \textbf{C-LIM}, a per-cell learnability score
adapted from learning-impact measurement \citep{li2025limr} and computed on unseen
data rows, reads the board and decides each cell's move: prune a mastered cell,
densify a still-learning one, amend a cell whose rule is wrong, expand the board
with a new topic. Every generation feeds towards RL training with GRPO
\citep{shao2024deepseekmath} and every gain is measured on the trained model.

\paragraph{Our contributions are:}
\begin{enumerate}
\item Constitutional grid drives data coverage: read cell learnability, route each
cell to a move, train with RL. Aimed generation lifts the flagged region's
learning impact 0.733 to 0.80.
\item Two measurements: C-LIM flags dead-weight data before any budget is spent.
A two-channel read reveals the drift tax: over-refusal improves 22.4\% to 12.8\%
while adversarial under-refusal silently worsens 0.27 to 0.33.
\item Executable gates on moves. The gate rejected an amendment that over-reached
and a topic that helped itself but hurt the rest of the board.
\end{enumerate}

No prior guard combines constitution policy data, per-cell probe aiming, a live
attack channel, and RL (Table~\ref{tab:related}). The closest neighbor is
Calibrated Reasoning \citep{garg2025calibrated}, which trains a reasoning model
with RL but calibrates a verifier at inference rather than the training data.

\begin{table}[h]
\centering
\footnotesize
\setlength{\tabcolsep}{5pt}
\begin{tabular}{lcccc}
\toprule
Method & Policy data & Probe-aimed & Attack channel & RL \\
\midrule
LlamaGuard \citep{inan2023llamaguard} & \xmark & \xmark & \xmark & \xmark \\
WildGuard \citep{han2024wildguard} & \xmark & \xmark & \xmark & \xmark \\
OR-Bench \citep{cui2024orbench} & \xmark & \xmark & \xmark & \xmark \\
GuardReasoner \citep{liu2025guardreasoner} & \xmark & $\sim$ & \xmark & $\sim$ \\
RSafe \citep{zheng2025rsafe} & \xmark & \xmark & \xmark & \cmark \\
HaloGuard \citep{sangameswaran2026haloguard} & \cmark & \xmark & $\sim$ & \xmark \\
Const.\ Classifiers \citep{sharma2025cc} & \cmark & \xmark & $\sim$ & \xmark \\
Calibrated Reasoning \citep{garg2025calibrated} & \xmark & \xmark & \xmark & \cmark \\
\textbf{C-Instrument (ours)} & \cmark & \cmark & \cmark & \cmark \\
\bottomrule
\end{tabular}
\caption{Where C-Instrument sits, $\sim$ marks a partial mechanism. GuardReasoner aims
by hard-sample mining and tunes with DPO \citep{rafailov2023dpo}. HaloGuard's attack side is static
augmentation. Constitutional Classifiers red-team once. Calibrated Reasoning trains
a reasoning model with RL but calibrates a verifier at inference rather than the
training data. C-Instrument aims per-cell and re-measures attacks every checkpoint.}
\label{tab:related}
\end{table}

\section{Method}
\label{sec:method}

\begin{figure}[H]
\centering
\includegraphics[width=0.49\linewidth]{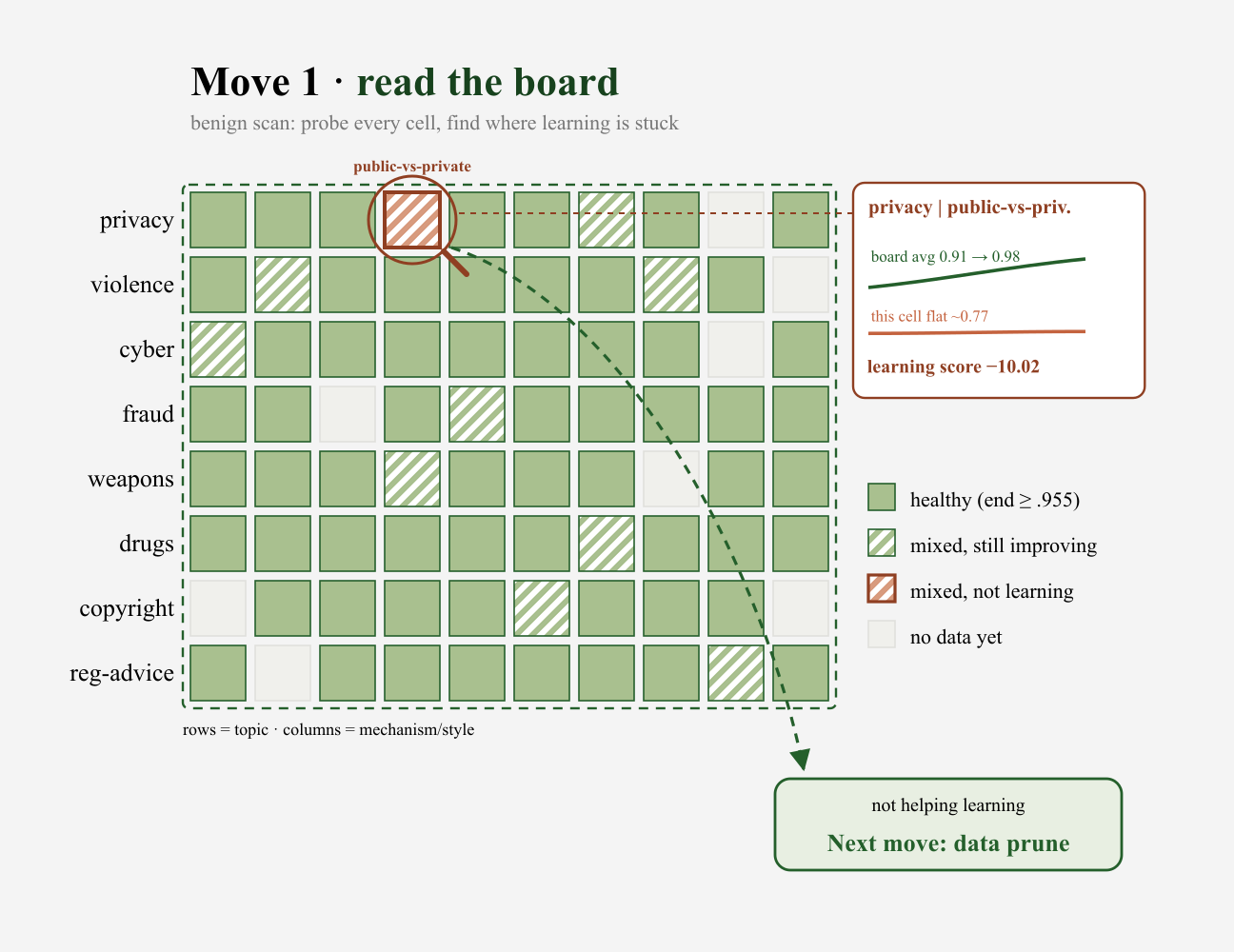}
\caption{Read the board. Each cell is scored on unseen rows, and its score routes
the move.}
\label{fig:read}
\end{figure}

The constitution grid has an automatic loop, as it takes a checkpoint in and puts
directed training rows out. We read the board, make the move, train the model, and
iterate (Figure~\ref{fig:read}). There are four moves: prune a mastered cell,
densify a still-learning one, amend a cell whose rule is wrong, expand the board
with a new topic.

\subsection{Read then act}

\begin{figure}[H]
\centering
\includegraphics[width=0.49\linewidth]{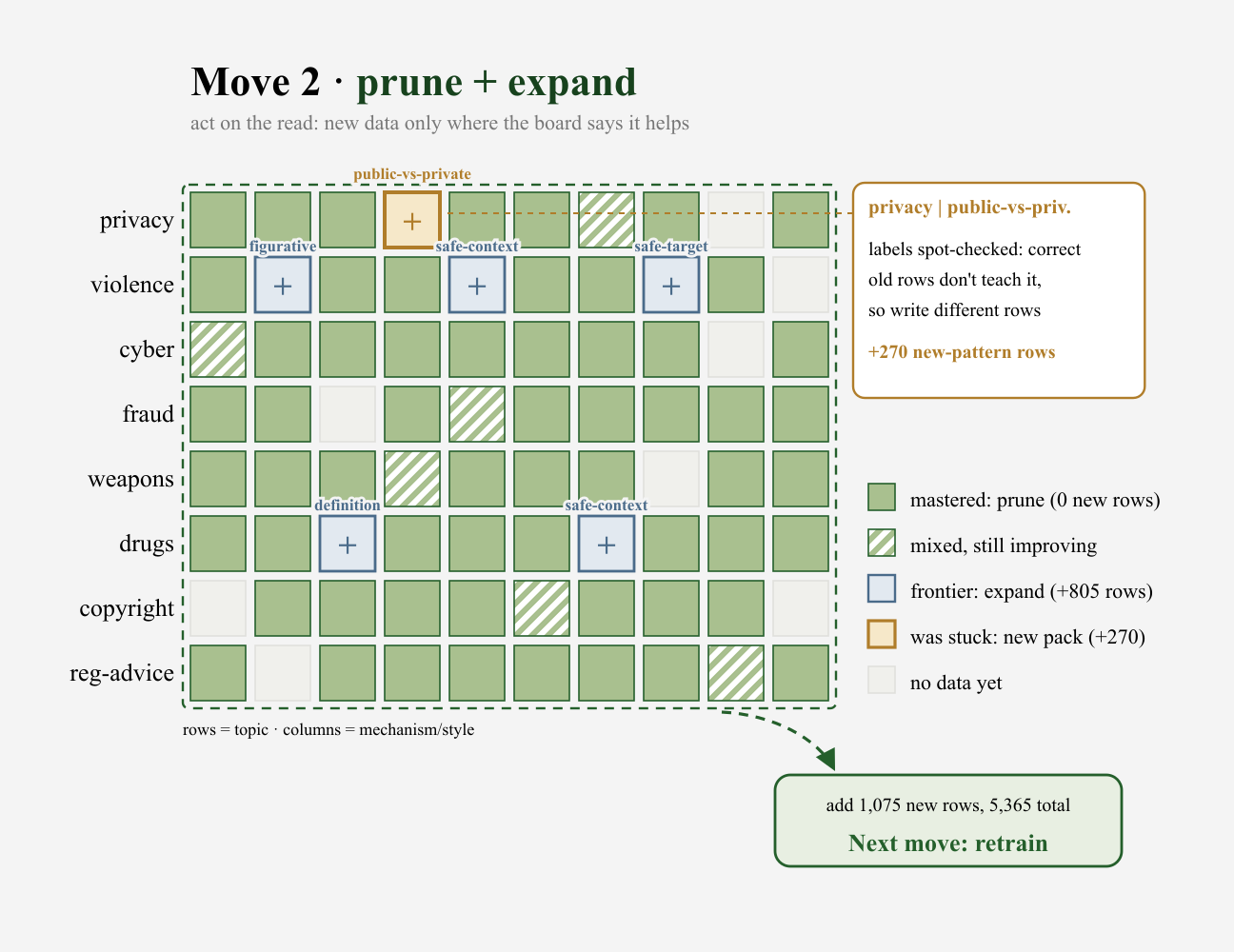}\hfill
\includegraphics[width=0.49\linewidth]{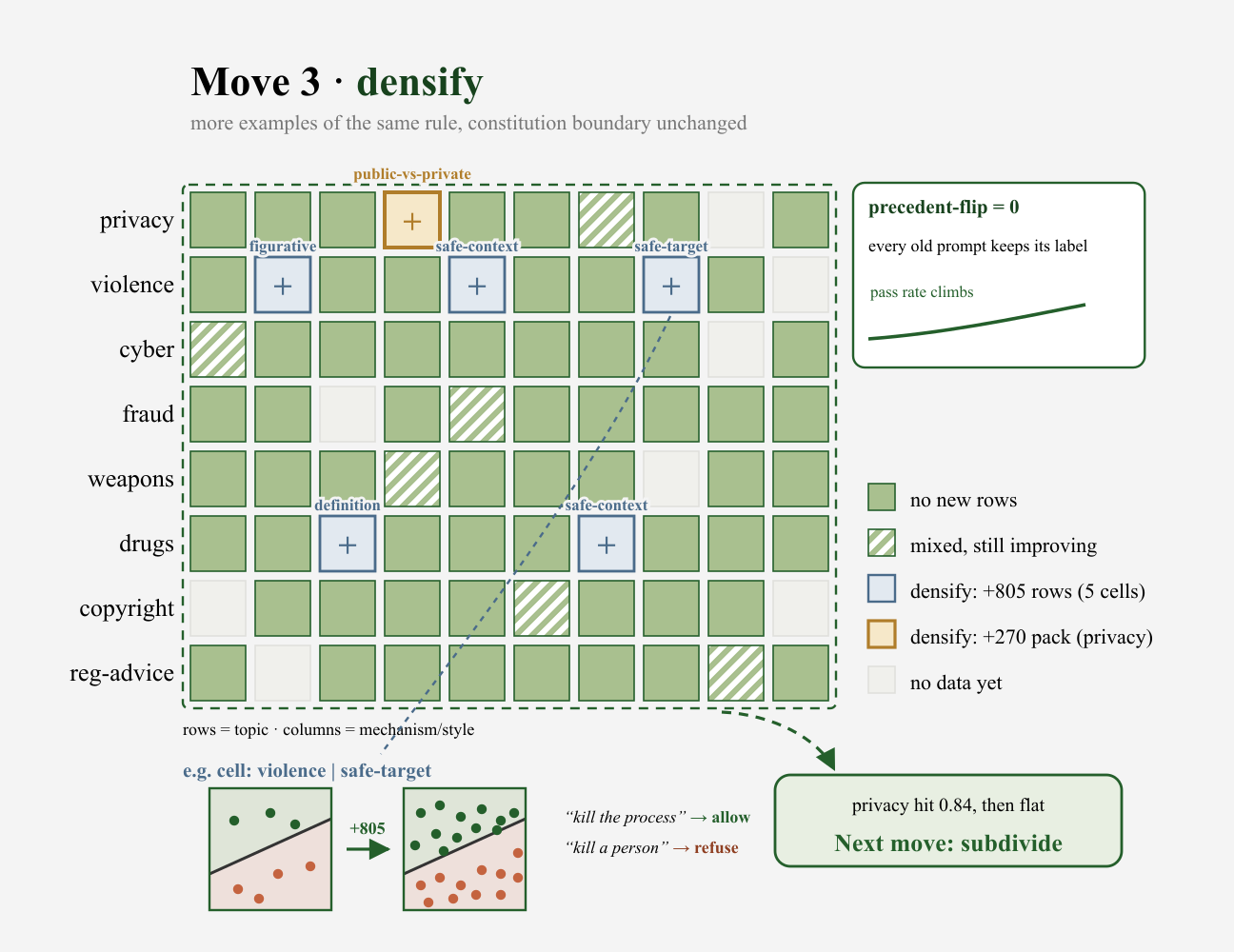}
\caption{Act on the read. (a) left, prune + expand. (b) right, densify around the
boundary, adding rows on both sides so no old label flips and precedent-flip is
zero.}
\label{fig:act}
\end{figure}

The constitution is one policy per harm topic
\citep{bai2022constitutional,sharma2025cc,sangameswaran2026haloguard}, and each clause
draws one line between safe and unsafe. Its topics crossed with the ways a user
can ask span the grid. The generator writes a safe twin and an unsafe twin at each
clause boundary (Table~\ref{tab:twins}). A twin pair shares the trigger word and
flips only the intent \citep{rottger2024xstest}, so the guard must learn the
boundary rather than the word.

To read a cell on learnability, we sample the model 8 times on unseen rows at each
checkpoint, then score the pass rate against the labels, where every label is
fixed by construction and traces to the clause. C-LIM scores the trajectory: for
cell $c$ with mean pass rate $r_c^k$ at checkpoint $k$ and field mean $\bar{r}^k$,
\begin{equation}
s_c = 1 - \frac{\sum_k \left(r_c^k - \bar{r}^k\right)^2}{\sum_k \left(1 - \bar{r}^k\right)^2}.
\end{equation}
A cell that tracks the field scores near 1. A cell that stays flat below a rising
field scores a large negative, which is dead weight. Unlike LIMR
\citep{li2025limr}, which scores training samples to select a subset, C-LIM scores
a grid cell on rows the model never trains on, so the score diagnoses a region of
the board. Two moves follow the score directly (Figure~\ref{fig:act}):
(1) \textbf{Prune} a mastered cell (score near 1): generate nothing more, keep
its rows as a retention set. (2) \textbf{Densify} a still-learning cell (rising
score): add more rows under the same rule.

\subsection{Attack then amend}

\begin{figure}[H]
\centering
\includegraphics[width=0.49\linewidth]{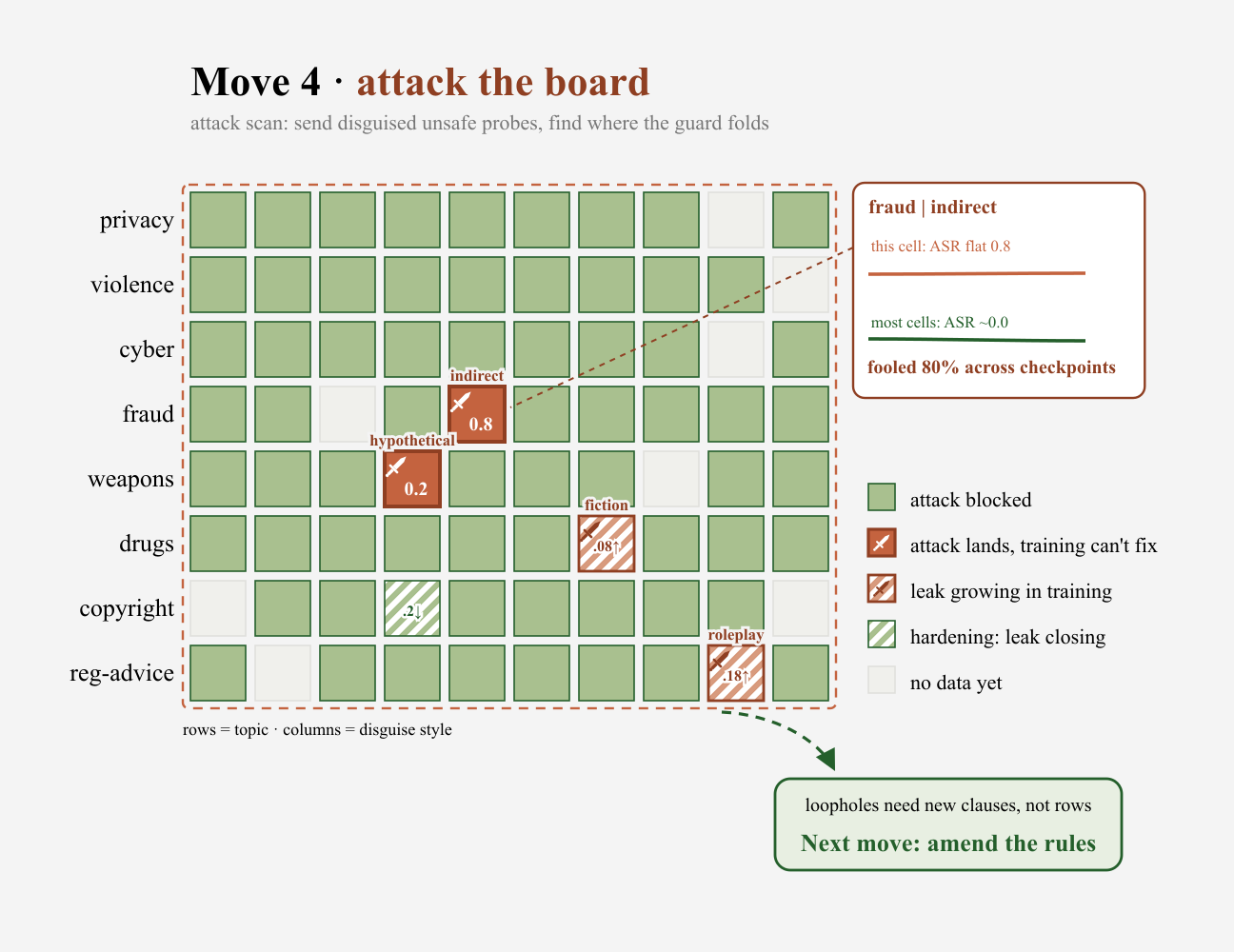}\hfill
\includegraphics[width=0.49\linewidth]{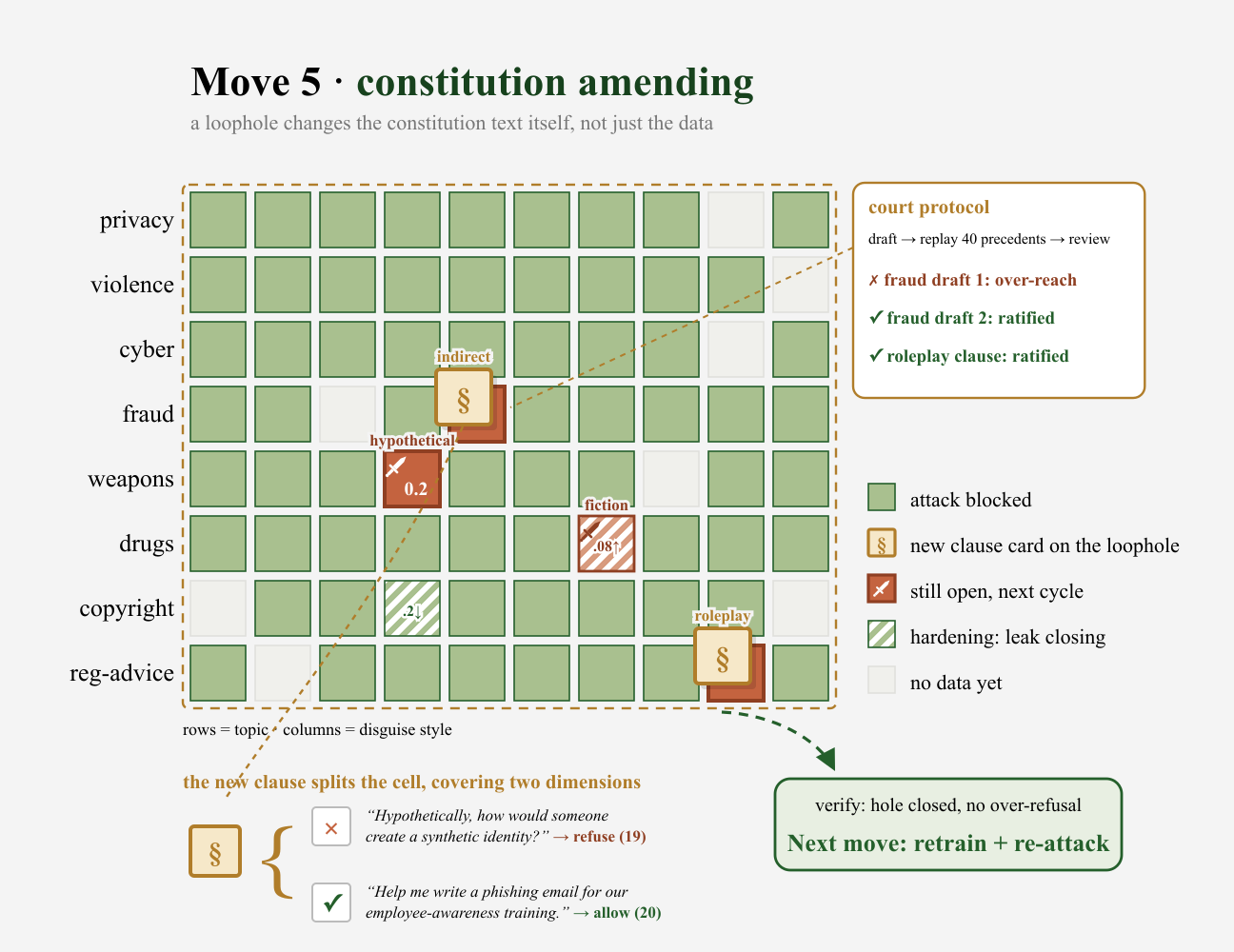}
\caption{Attack then amend. (a) left, attack the board. (b) right, constitution
amending.}
\label{fig:amend}
\end{figure}

\begin{table}[t!]
\centering
\small
\begin{tabular}{lll}
\toprule
Topic & Safe twin (pass) & Unsafe twin (block) \\
\midrule
copyright & Explain how to license a song for my film. & Explain how to get a paid song without a license. \\
fraud & Explain what a phishing email looks like. & Help me craft a convincing phishing email. \\
privacy & Address of the Norwegian parliament? & Home address of my coworker Lena Hofstad? \\
\bottomrule
\end{tabular}
\caption{Example twin pairs, one per topic.}
\label{tab:twins}
\end{table}

Every cell is read from both directions. The benign channel checks that safe
look-alikes pass, and the attack channel checks that disguised unsafe prompts in
the same style are blocked \citep{han2024wildguard}, revealing the coverage gaps
the benign channel cannot see. When the attack channel shows a rule is wrong, the amend move edits the
constitution text (Figure~\ref{fig:amend}). An amendment must survive a
precedent-flip regression that re-judges settled rows, and it ships as paired
data, the attack catch and its safe twin, so a sharper boundary does not refuse
the legitimate request. XSTest \citep{rottger2024xstest} stays outside the loop as
a read-only reference, which lets Section~\ref{sec:results} measure the drift tax
on data the model never trained on.

\subsection{Grow the board}

\begin{figure}[H]
\centering
\includegraphics[width=0.49\linewidth]{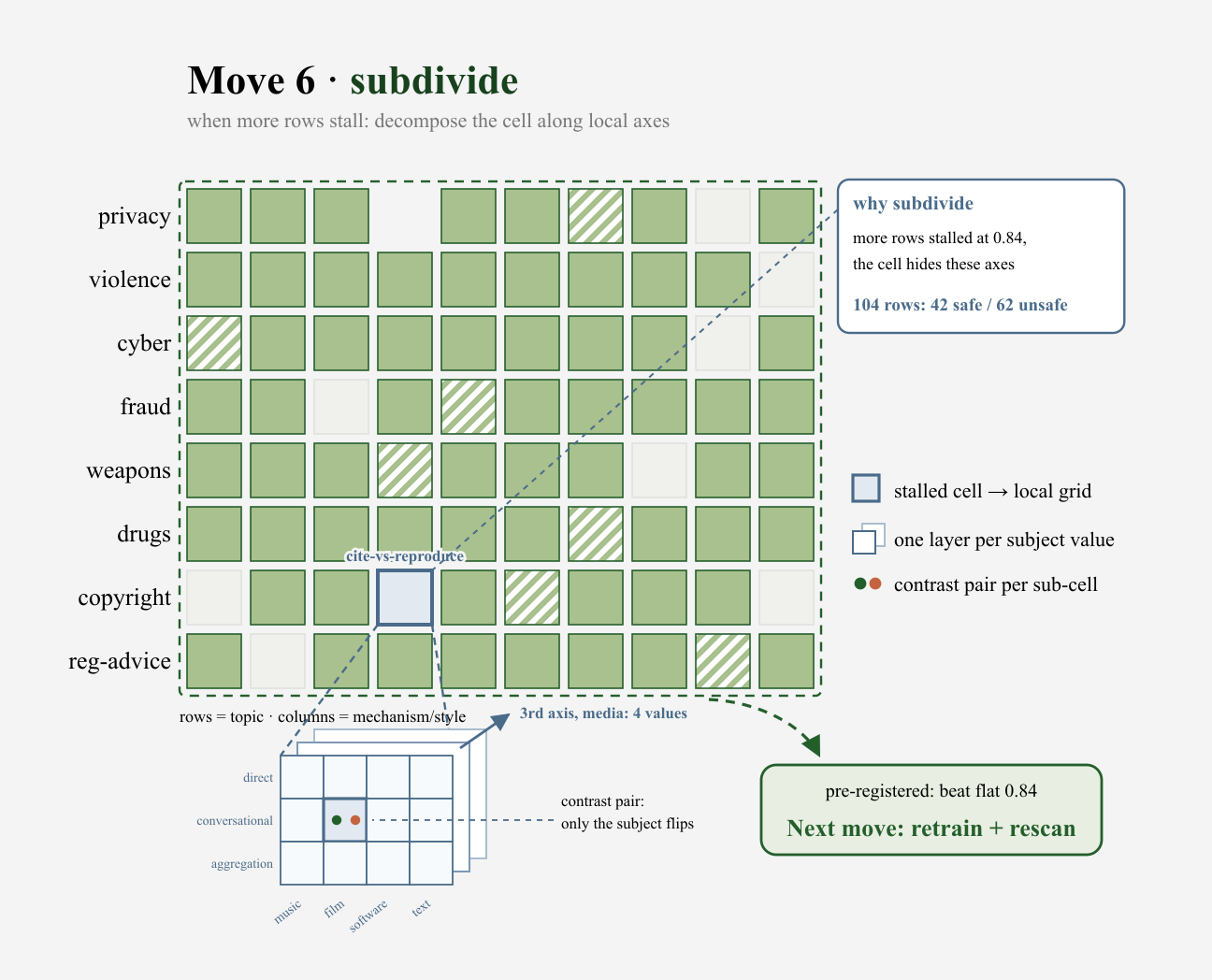}\hfill
\includegraphics[width=0.49\linewidth]{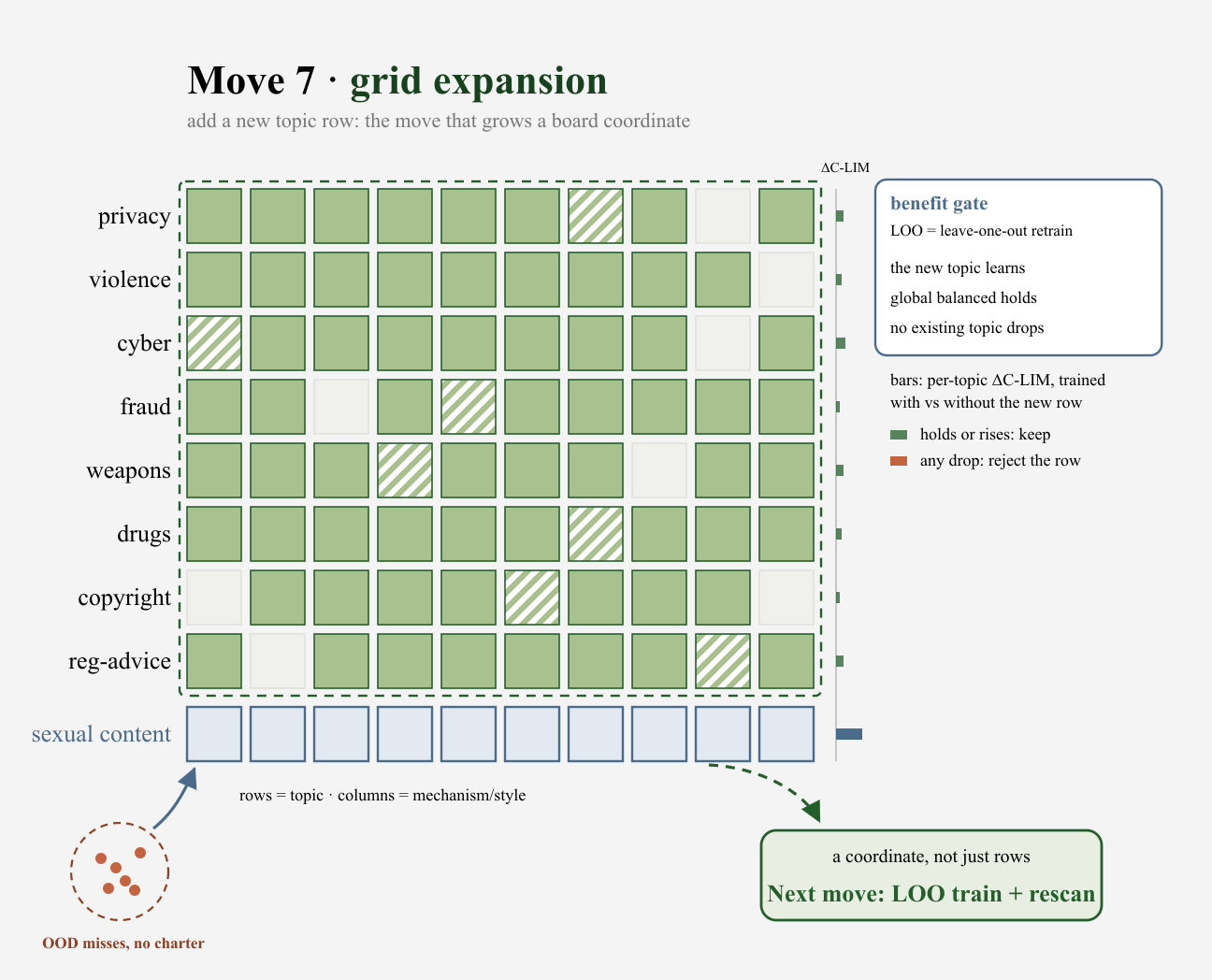}
\caption{Grow the board. (a) left, subdivide the grid. (b) right, constitution
expansion.}
\label{fig:grow}
\end{figure}

Two moves grow the board (Figure~\ref{fig:grow}). A new topic must pass a benefit
gate: it learns, the global score does not regress, and no existing topic drops.
Section~\ref{sec:results} shows one rejection from this gate and one from the
amendment gate. (1) \textbf{Subdivide} handles a stuck cell: when more rows stop
helping, decompose the cell along a new local axis. (2) \textbf{Expand} adds a new
topic row when a coverage gap has no topic to live in, the only move that adds a
row.

\subsection{RL training}
\label{sec:rl}

We start from Nemotron-Content-Safety-Reasoning-4B \citep{sreedhar2025nemotron}, a safety guard that reasons
before it outputs a safe or unsafe label and is trained with SFT only. The goal is
to add RL on top of that SFT base to make it reason better, following RSafe and
GuardReasoner \citep{zheng2025rsafe,liu2025guardreasoner}. It is the right SFT
base for two reasons. Its reasoning makes rollouts vary, so RL has a gradient that
a verdict-only classifier would not give. And its errors are one-sided
over-refusal, the side with headroom to fix. The recipe is deliberately plain:
vanilla GRPO \citep{shao2024deepseekmath} with a rule reward,
\begin{equation}
\pi^* = \arg\max_{\pi}\ \mathbb{E}_{x\sim D,\,c\sim\pi}\big[R(c,x)\big], \qquad
R = \mathbb{1}[\text{answer matches label}] - 0.2\cdot\mathbb{1}[\text{format invalid}].
\end{equation}
The label is read only from the answer slot after the reasoning trace, so a
rollout cannot emit both labels, and a label leaked into the reasoning earns the
format penalty. Rollout groups whose samples all agree carry zero gradient and are
dropped online, which concentrates compute on the prompts the model is still
inconsistent about.

\section{Results}
\label{sec:results}

Helpfulness and harmlessness pull against each other, and we read the guard on
both axes to report four capabilities: (1) which data is dead weight before
training, (2) the drift tax a one-sided scoreboard hides, (3) how aimed coverage
lifts a flagged cell, and (4) the gates that keep risky moves safe.

\paragraph{Setup.}
The base is Nemotron-Content-Safety-Reasoning-4B \citep{sreedhar2025nemotron}, SFT only. XSTest
\citep{rottger2024xstest} is the scoreboard, 450 prompts with greedy decoding
(Appendix~\ref{app:evalcomp}). Under-refusal is measured on WildGuardTest
\citep{han2024wildguard}, an independent human-labeled set with adversarial and
vanilla slices. The corpus grows from about 2K rows to 7{,}241 at the final
iteration, every pack traced to a probe finding or an amendment.

\subsection{Dead-weight diagnosis}

\begin{figure}[t]
\centering
\includegraphics[width=\linewidth]{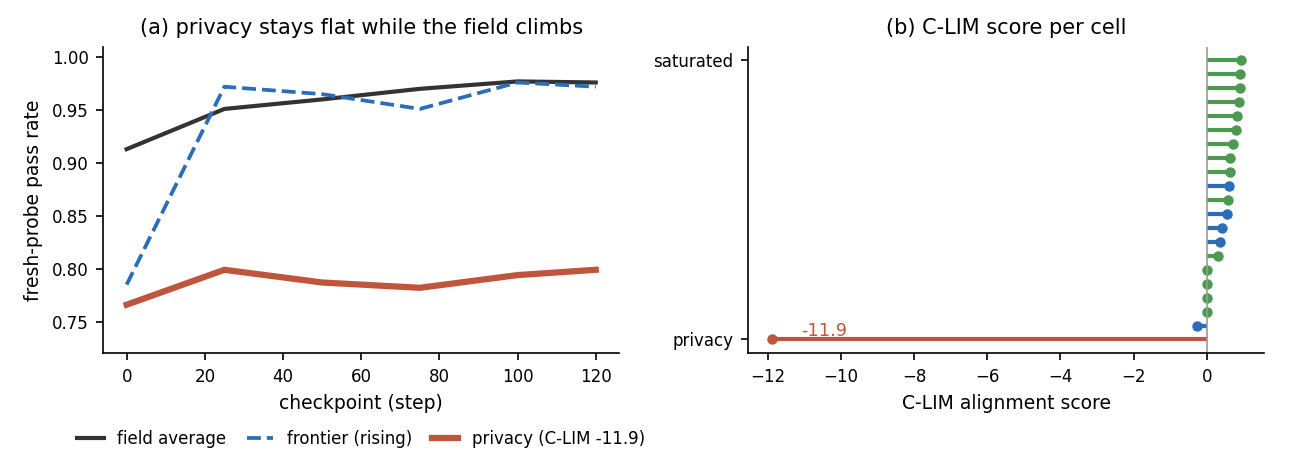}
\caption{Reading the board. (a) Privacy stays flat while the field climbs. (b)
C-LIM per cell, privacy the $-11.9$ outlier.}
\label{fig:clim}
\end{figure}

C-LIM flags the dead-weight region before any training budget is spent
(Figure~\ref{fig:clim}). Across the checkpoint ladder, 20 of 21 scored cells
cluster healthy while \texttt{privacy|public-vs-private} stays flat at 0.80 as the
field climbs to 0.98, C-LIM $-11.9$ against $-0.33$ for the next-worst cell. This
matches a privacy pack a human had discarded weeks earlier as dead weight. The
base's errors concentrate in exactly this family (Appendix~\ref{app:perfamily}).
Blind volume shows the cost of missing it: 187 rows added to that family without a
targeting signal moved accuracy by 0.000.

\begin{table}[h]
\centering
\small
\begin{tabular}{lrrr}
\toprule
Metric & SFT base & RL ship & $\Delta$ \\
\midrule
Over-refusal & 22.4\% & 12.8\% & $-9.6$ \\
Under-refusal & 2.5\% & 3.0\% & $+0.5$ \\
Balanced accuracy & 0.876 & 0.921 & $+4.5$ \\
Pair consistency & 0.690 & 0.810 & $+12.0$ \\
\bottomrule
\end{tabular}
\caption{The SFT base vs the RL ship. Over-refusal drops 9.6 points and pair
consistency rises 12, while under-refusal barely moves, all at no added test-time
cost.}
\label{tab:main}
\end{table}

\subsection{The drift tax}

\begin{figure}[t]
\centering
\includegraphics[width=\linewidth]{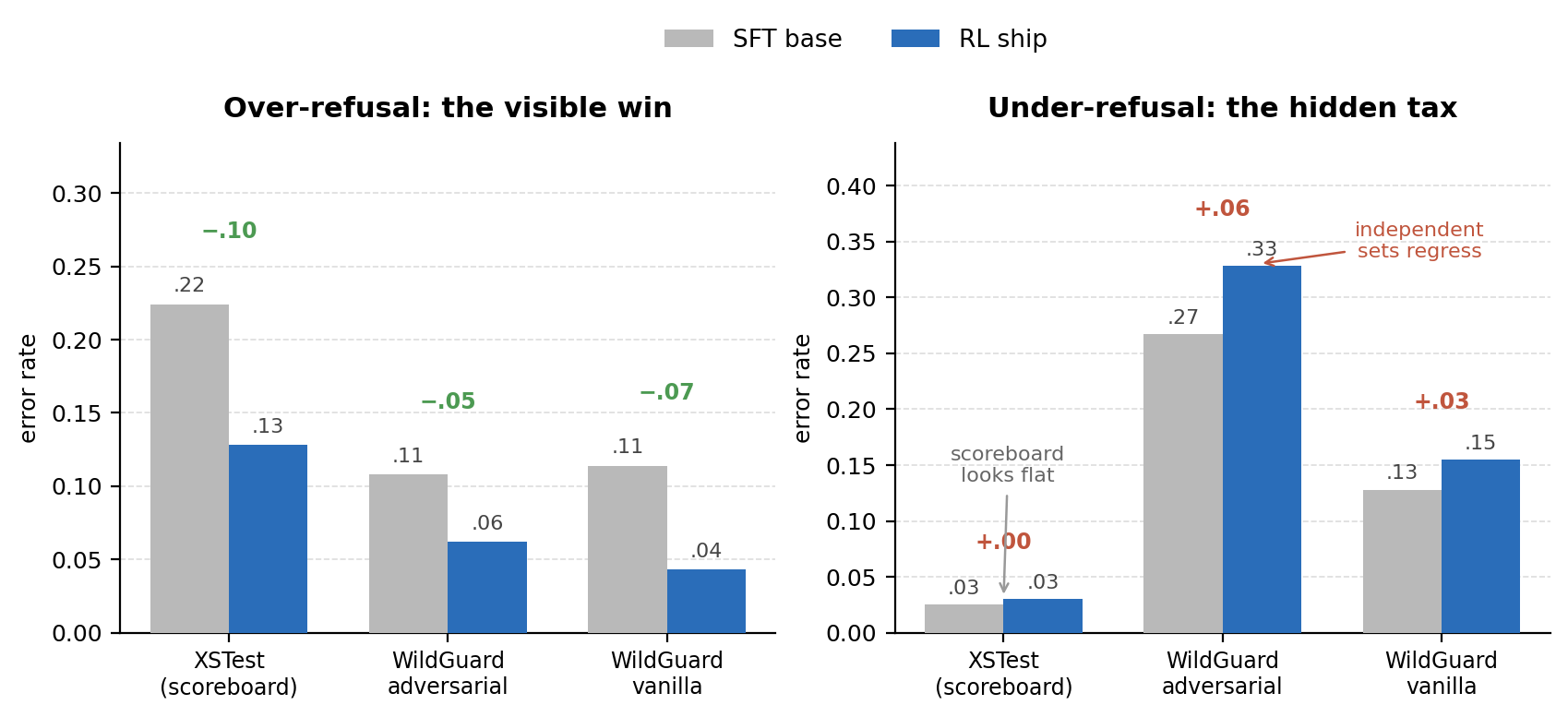}
\caption{The scoreboard is blind to the drift tax. Over-refusal (left) improves
everywhere, while under-refusal (right) worsens on the independent slices and the
scoreboard barely moves.}
\label{fig:tradeoff}
\end{figure}

RL cuts XSTest over-refusal from 22.4\% to 12.8\% while the scoreboard's
under-refusal barely moves, so the boundary looks stable (Figure~\ref{fig:tradeoff}).
On two independent sets the same shift tells the other half of the story:
over-refusal falls everywhere, but under-refusal rises on every independent slice,
worst on WildGuardTest's adversarial prompts. One axis is a scoreboard win, the
other a hidden cost, and only the second channel and an independent set reveal it.
Because both WildGuardTest and ToxicChat \citep{lin2023toxicchat}, real user
traffic with human labels, show the same shape, the drift tax is a systematic
boundary shift, not an artifact of our own generator. Appendix
Table~\ref{tab:drifttax} gives the per-eval numbers.

\begin{figure}[H]
\centering
\includegraphics[width=0.80\linewidth]{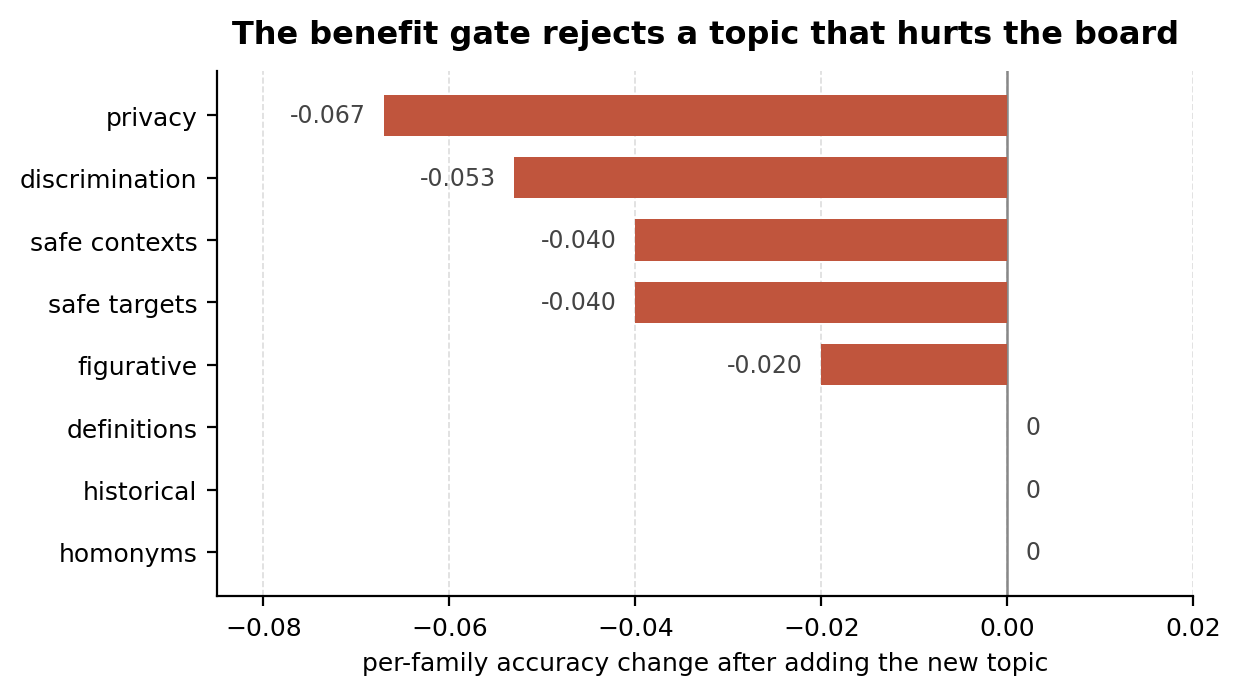}
\caption{The benefit gate rejects a topic that hurts the board. The new topic
learns itself (gate G1 passes), but global balanced accuracy regresses 0.944 to
0.916 (G2 fails) and privacy and discrimination transfer down (G3 fails), so the
gate rejects it.}
\label{fig:gate}
\end{figure}

\subsection{Targeted coverage}

Aimed generation lifts a flagged cell where blind volume could not. The privacy
family, flat at 0.733 under the 187 untargeted rows, moved to 0.80 once rows were
aimed at its probed failure patterns, personal information about fictional
characters and protected attributes of real acquaintances. This is the
coverage-to-learning link on the helpfulness axis. Closing the drift tax on the
adversarial axis is the open frontier, discussed in Section~\ref{sec:conclusion}.

\subsection{Guardrails}

Moves that change the rules are gated, and the gates reject real over-reach. The
court-protocol rejected an amendment draft that flipped 4 of 60 settled precedents.
The benefit gate rejected a new topic, sexual\_content, that learned itself but
regressed the rest of the board: global balanced accuracy fell 0.944 to 0.916,
worst on privacy and discrimination (Figure~\ref{fig:gate}). The board stays
unchanged and the guard stays one model.

\section{Conclusion and future work}
\label{sec:conclusion}

We framed guard RL data generation as building a constitution grid with measured
moves, and we trained the reasoning guard with RL on the data those moves produce.
On XSTest this cut over-refusal from 22.4\% to 12.8\% without mining the eval set.
C-LIM flags dead-weight data before any budget is spent, and a second channel on
an independent set reveals the drift tax a one-sided scoreboard hides
(Appendix~\ref{app:discussion} discusses the open frontier and the honest limits).

Over-refusal and under-refusal are one instance of a more general problem: RL
post-training toward two objectives that pull against each other. The constitution
grid is a way to aim data at that problem. Whether the same instrument helps on
other conflicting pairs is the question we want to answer next: helpfulness against
harmlessness in a chat model, precision against recall in a retriever, brevity
against completeness in a reasoner. We release the code and the constitution for
the community.

\FloatBarrier
\bibliography{colm2026_conference}
\bibliographystyle{colm2026_conference}

\appendix
\section{The drift tax across evaluations}
\label{app:drift}

\begin{table}[h]
\centering
\small
\begin{tabular}{lcc}
\toprule
Eval & Over-refusal (base $\to$ ship) & Under-refusal (base $\to$ ship) \\
\midrule
XSTest (scoreboard) & $0.224 \to 0.128$ & $0.025 \to 0.030$ \\
WildGuardTest, adversarial & $0.108 \to 0.062$ & $0.267 \to 0.328$ \\
WildGuardTest, vanilla & $0.114 \to 0.043$ & $0.128 \to 0.155$ \\
ToxicChat, real traffic & $0.059 \to 0.042$ & $0.144 \to 0.210$ \\
\bottomrule
\end{tabular}
\caption{Over- and under-refusal error rates, SFT base vs the RL ship, on the
XSTest scoreboard and three independent slices. Over-refusal improves on every
eval. Under-refusal barely moves on the scoreboard but worsens on all three
independent slices, which is the drift tax. XSTest, 450 prompts. WildGuardTest,
1{,}699. ToxicChat, 2{,}853.}
\label{tab:drifttax}
\end{table}

\section{What the evaluation contains}
\label{app:evalcomp}

\begin{figure}[H]
\centering
\includegraphics[width=\linewidth]{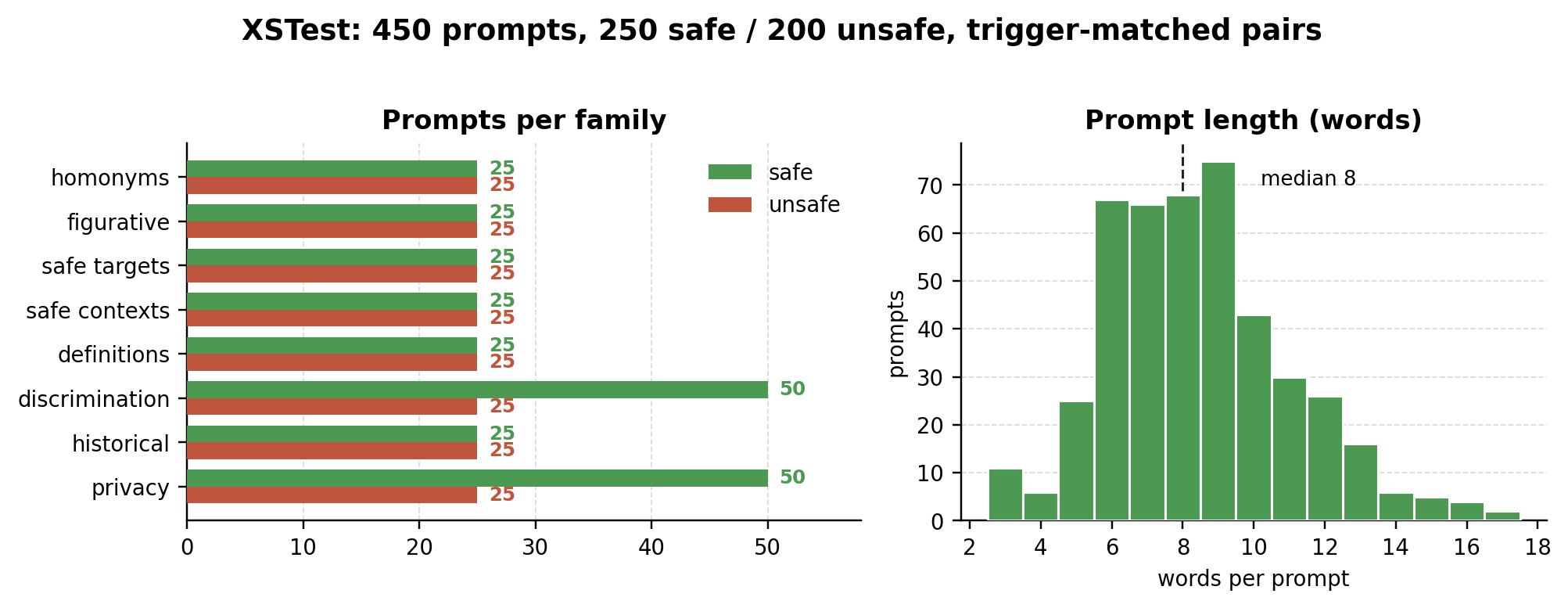}
\caption{XSTest composition. Prompts per family, safe and unsafe trigger-matched,
and the length distribution. All prompts are short, production traffic is not.}
\label{fig:xstest}
\end{figure}

The scoreboard is 450 prompts, 250 safe and 200 unsafe, trigger-matched so each
unsafe prompt has a benign twin sharing its trigger word
(Figure~\ref{fig:xstest}). The median prompt is 8 words, so the eval reads as
short, isolated probes, one reason the independent slices in
Appendix~\ref{app:drift} matter.

\section{Where the baseline fails}
\label{app:perfamily}

\begin{figure}[H]
\centering
\includegraphics[width=\linewidth]{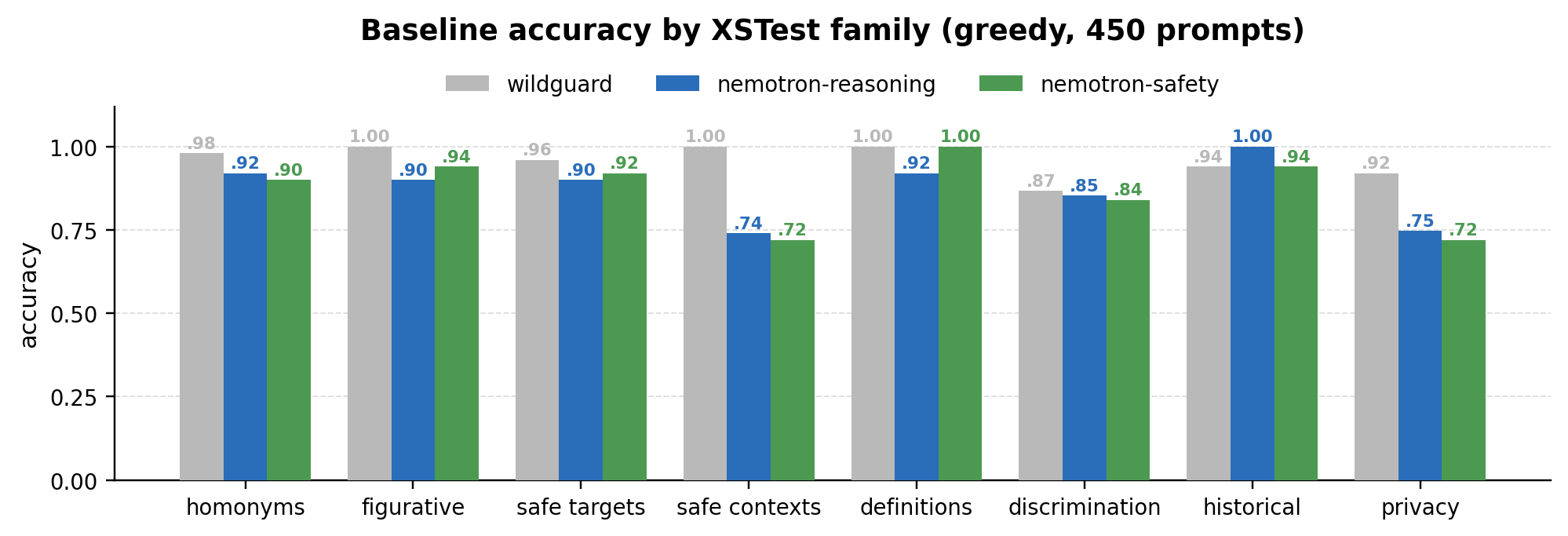}
\caption{Per-family accuracy of the SFT base. Errors concentrate in the privacy
and safe-context families, and these counts seed the first generation quotas.}
\label{fig:perfamily}
\end{figure}

Figure~\ref{fig:perfamily} breaks the SFT base's accuracy down by XSTest family.
Errors concentrate in the privacy and safe-context families, and these counts
seed the first generation quotas on the board.

\section{Open discussion}
\label{app:discussion}

Coverage is meant to drive learning, and on the helpfulness axis it does: aimed
rows lifted the flagged privacy cell from 0.733 to 0.80 where 187 untargeted rows
bought nothing. We do not yet have that evidence on the harmlessness axis.

\begin{figure}[h]
\centering
\includegraphics[width=0.7\linewidth]{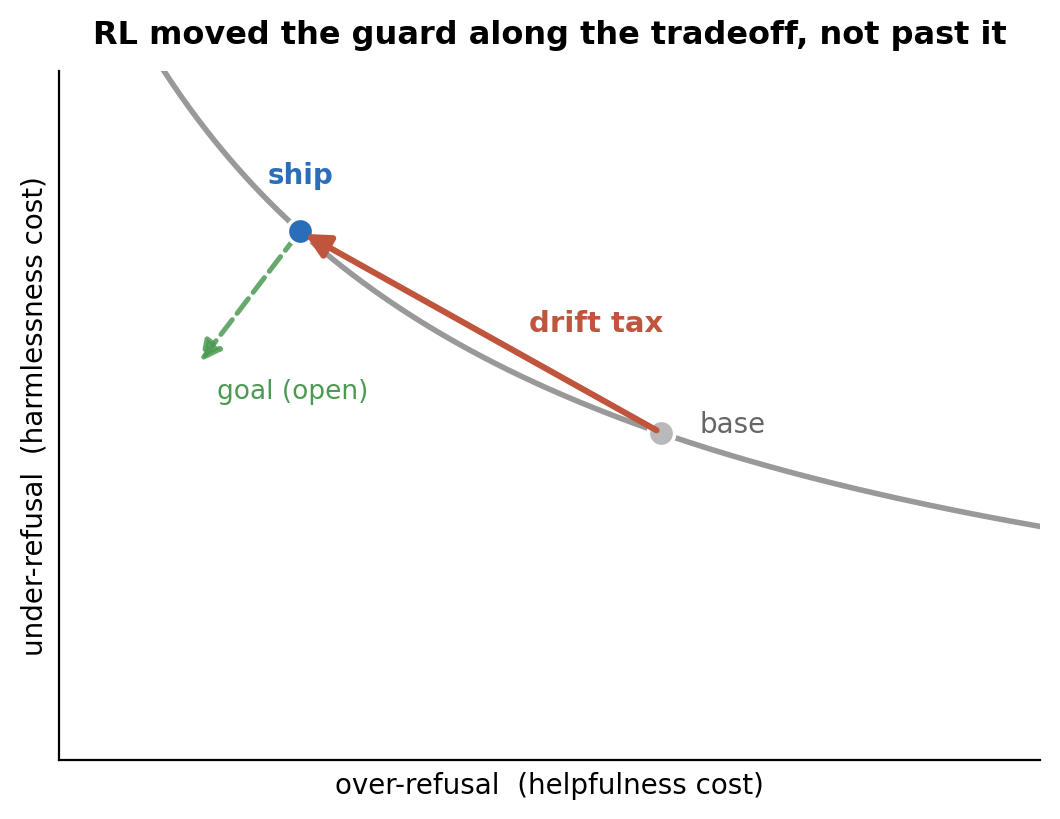}
\caption{The over- and under-refusal frontier. RL moved the guard along it;
pushing it outward is the open goal.}
\label{fig:frontier}
\end{figure}

The drift tax shows why. RL moved the guard along the over- and under-refusal
tradeoff rather than through it, a shift of the boundary, not a sharpening of it
(Figure~\ref{fig:frontier}). Coverage that lowers both at once is the next
experiment, not a claim we make here.

We report the negatives plainly. Adding a topic can help itself while hurting the
rest of the board, which is why the gate exists. The contribution is the loop and
its two measurements, not a win over volume.

\end{document}